\documentclass[lettersize,journal]{IEEEtran}

\usepackage[caption=false,font=normalsize,labelfont=sf,textfont=sf]{subfig}
\usepackage{amsmath}
\usepackage{amssymb}
\usepackage{amsthm}
\usepackage{amsfonts}
\usepackage{mathrsfs}
\usepackage{mathtools}
\usepackage{array}
\usepackage{algorithm}
\usepackage{algorithmic}
\usepackage{graphicx}
\usepackage{textcomp}
\usepackage{xcolor}

\usepackage{cite}
\usepackage{stfloats}
\usepackage{verbatim}
\usepackage[utf8]{inputenc} 
\usepackage[T1]{fontenc}    
\usepackage{hyperref}       
\usepackage{url}            
\usepackage{booktabs}       
\usepackage{nicefrac}       
\usepackage{microtype}      
\usepackage{latex_style_karim2022}

\def\MatrSigma{\bogs{\Sigma}}

\def\CV{\textit{CV}}

\def\smallACVL{\textit{acvl}}
\def\smallCWH{\textit{cwh}}
\def\smallZW{\textit{zw}}

\newcommand{\indep}{\perp \!\!\! \perp}

\newtheorem{proposition}{Proposition}[section]

\begin{document}


\title{Efficient Estimation of Regularized Tyler's M--Estimator Using Approximate LOOCV}

\author{Karim Abou--Moustafa,~\IEEEmembership{Senior Member,~IEEE}
\thanks{Karim Abou--Moustafa is with Intel's Foundry Technology Development (TD) Division, Intel Corp., Chandler, Arizona, USA. 
Email: \texttt{Karim.Abou-Moustafa@intel.com}; \texttt{karim.aboumoustafa@gmail.com}}
}

\maketitle


\begin{abstract}
We consider the problem of estimating a regularization parameter, or a shrinkage coefficient $\alpha \in (0,1)$ for 
Regularized Tyler's M-estimator (RTME).
In particular, we propose to estimate an optimal shrinkage coefficient by setting $\alpha$ as the solution to a suitably 
chosen objective function; namely the leave-one-out cross-validated (LOOCV) log-likelihood loss.
Since LOOCV is computationally prohibitive even for moderate sample size $n$, we propose a computationally efficient 
approximation for the LOOCV log-likelihood loss that eliminates the need for invoking the RTME procedure $n$ times for 
each sample left out during the LOOCV procedure.
This approximation yields an $O(n)$ reduction in the running time complexity for the LOOCV procedure, which results 
in a significant speedup for computing the LOOCV estimate. 
We demonstrate the efficiency and accuracy of the proposed approach on synthetic high-dimensional data sampled from 
heavy-tailed elliptical distributions, as well as on real high-dimensional datasets for object recognition, face 
recognition, and handwritten digit's recognition. 
Our experiments show that the proposed approach is efficient and consistently more accurate than other methods in the 
literature for shrinkage coefficient estimation.
\end{abstract}




\section{Introduction}
\label{sec:intro}

\IEEEPARstart{C}{ovariance} matrices, or their scaled versions \emph{scatter matrices}, are ubiquitous in statistical models 
and procedures for machine learning, pattern recognition, signal processing, and various other fields of 
science and engineering.
The performance of procedures such as 
principal component analysis (PCA) and its extensions 
\cite{book_pca_jolliffe_2002}, 
linear discriminant analysis (LDA) and its extensions 
\cite{fisher_da_36,karim_delatorre_ferrie_PRJ2015}, 
canonical correlation analysis (CCA) \cite{hotelling_cca_1936},
portfolio optimization for investment diversification \cite{markowitz_1952},
outlier detection using robust Mahalanobis distance \cite{book_rouss_87,cabana_lillo_laniado_2019}, and
covariance descriptors \cite{tuzel_porikli_meer_2006},
all depend on an accurate estimate of the covariance matrix.
Unfortunately, the process of accurately estimating a covariance matrix is challenging since the number of 
unknown parameters grows quadratically with the data dimensionality $p$.
The problem is well-understood when the number of samples $n$ is much larger than $p$ and the data's 
underlying distribution is a multivariate Gaussian.
In this case, the sample covariance matrix (SCM) is an accurate estimate of the covariance matrix, and 
is optimal under most criteria \cite{wiesel_2012}.
In various modern applications, however, $p$ may be comparable to, or greater than $n$, and the data's 
underlying distribution may be non-Gaussian and/or \emph{heavy-tailed}.
The situation gets exacerbated if the data are also contaminated with outliers.
In such settings, the SCM is known to be a poor estimate of the covariance matrix and one needs to consider 
estimators that are more accurate and robust than the SCM.
In this work, we are interested in a particular estimator from the family of \emph{robust} and \emph{affine-invariant}
$M$-estimators of scatter matrices proposed by Marona \cite{marona_robust_mestimators_1976}  
-- namely Tyler's $M$-estimator \cite{tyler_tme_1987,tyler_acg_tme_1987} -- in the setting where the data's distribution 
is \emph{heavy-tailed} and the sample support is relatively low; i.e. the number of variables (features) $p$ is large 
and $p \geq n$.\footnote{See \cite{marona_yohai_2017,couillet_pascal_silverstein_2014,wiesel_zhang_book_2015} 
for a recent overview and results on this family of estimators.}

Various approaches were proposed for estimating high-dimensional covariance matrices when $p \geq n$;
shrinkage-based approaches \cite{james_stein_1961,dey_srinivasan_1985,daniels_kass_2001,ledoit_wolf_2004};
specifying an appropriate prior distribution for the covariance matrix \cite{haff_1980};
regularization-based approaches \cite{daspremont_elghaoui_2008,ravikumar_wainright_yu_2011,book_poorahmadi_2013}; 
approaches that exploit sparsity assumptions (banding, tapering, thresholding) 
\cite{bickel_levina_covreg_thresh_2008,elkaroui_2008,cai_zhou_minimax_est_large_cov_2012,han_liu_2017}; and 
approaches developed in the robust statistics literature \cite{book_huber_81,book_hampel_stahel_2013}.
With the exception of some approaches from the robust statistics literature, most of the other approaches assume 
that the data's underlying distribution is a multivariate Gaussian which may not be suitable for handling outliers, 
or samples from heavy-tailed distributions.

Tyler's $M$-estimator (TME) is an accurate and efficient robust estimator for the scatter matrix when the data 
are samples from an \emph{elliptical distribution} with heavy-tails and $n \gg p$. 
Elliptical distributions (introduced shortly) are the generalization of the multivariate Gaussian distribution 
and are suitable for modelling empirical distributions with heavy tails, where such heavy tails may be due to the 
existence of outliers in the data \cite{kelker_1970,cambanis_1981}.
In this setting, and under some mild assumptions on the data, TME has various attractive properties 
\cite{tyler_tme_1987,tyler_acg_tme_1987}. 
In particular, TME is strongly consistent, asymptotically normal, and is the \emph{most robust} estimator for the 
scatter matrix for an elliptical distribution in a \emph{minimax} sense; minimizing the maximum asymptotic variance 
(see Remark 3.1 in \cite{tyler_tme_1987}).  
Unfortunately in the $p > n$ regime, Tyler's $M$-estimator is not defined.
Various research works have proposed regularized versions of TME using the spirit of Ledoit \& Wolf \cite{ledoit_wolf_2004} 
linear shrinkage estimator model whose performance depends on a \emph{carefully chosen} regularization parameter, 
or \emph{shrinkage coefficient} $\alpha \in (0,1)$
\cite{abramovich_spencer_2007,chen_wiesel_hero_2011,wiesel_2012,pascal_chitour_queck_2014,sun_babu_palomar_2014,ollila_tyler_2014,zhang_wiesel_2016,abramovich_besson_2_2013}.
Our work here addresses the question of shrinkage coefficient estimation for \emph{Regularized} TME (RTME), 
and proposes a computationally efficient algorithm for obtaining a near-optimal estimate for this parameter.\footnote{Shrinkage 
coefficient estimation for SCM and \emph{generalized} $M$-estimators for elliptically distributed data
was considered in \cite{ollila_raninen_2019,ollila_palomar_pascal_2021,ashurbekova_2021}.\\}

Unfortunately, the recursive nature of TME's procedure makes estimating an optimal shrinkage coefficient 
for this estimator a non-trivial problem.
Arguably, three broad approaches were considered to address this problem: 
(\emph{i}) oracle and random matrix theory (RMT) based approaches 
\cite{chen_wiesel_hero_2011,ollila_tyler_2014,couillet_mckay_2014,zhang_wiesel_2016,hoarau_2017,ashurbekova_2021};
(\emph{ii}) data-dependent approaches based on \emph{Cross Validation} (CV) techniques 
\cite{abramovich_spencer_2007,wiesel_2012,sun_babu_palomar_2014,dumbgen_tyler_2016};
and 
(\emph{iii}) maximum likelihood based approaches \cite{abramovich_besson_2_2013}.

Oracle-based approaches are computationally efficient due their closed-form solutions but may come short 
in terms of accuracy due to their implicit assumptions on the data distribution, and the implicit assumptions 
in their asymptotic estimates. 
CV techniques on the other hand are more accurate than oracle based methods since they are data-dependent 
approaches; this accuracy, however, comes at the cost of intensive computations, especially for high-dimensional 
data, which makes CV techniques not a favorable option for various applications.
Last, the maximum likelihood (ML) approach was considered in \cite{abramovich_besson_2_2013} where the Authors 
develop an approach, namely the expected likelihood (EL) method, for selecting a shrinkage coefficient for RTME 
when used for some specific problems in wireless communications;  
\emph{e.g.} adaptive-filtering and estimating the signal's direction of arrival.
While in such applications the noisy data samples may be reasonably assumed to have an elliptical distribution, 
the EL method may not be considered a general approach for estimating the shrinkage coefficient due to the specialized 
and controlled environments for such problems in wireless communications.

In this paper, we propose a more general approach for estimating an optimal shrinkage coefficient $\alpha^*$ for RTME.
Our proposed approach formulates the problem of estimating $\alpha^*$ as an optimization problem with respect to parameter $\alpha$.
In particular, we define an optimal shrinkage coefficient $\alpha^*$ as the minimizer for the following loss function;
the leave-one-out cross-validated (LOOCV) negative log-likelihood (NLL) for the estimated scatter matrix with respected 
to parameter $\alpha$ (Eq. \ref{eq:opt_alpha_exact_loocv}).
Since LOOCV scales linearly with the number of samples $n$ and hence is computationally prohibitive, we propose a 
computationally efficient \emph{approximation} for the LOOCV NLL loss function that \emph{eliminates} the need for 
computing the Regularized TME $n$ times for each sample left out during the LOOCV procedure.
The proposed approximation leverages the asymptotic properties of LOOCV estimates under a suitable notion of 
algorithmic stability.
This approximation yields an $O(n)$ reduction in the running time complexity for the LOOCV procedure, which results in 
a significant speedup in computing the LOOCV NLL loss.

At a high-level, the resulting procedure, namely the Approximate Cross-Validate Likelihood (ACVL) method, exploits 
mild computation and the given finite sample to select a (\emph{data-dependent}) \emph{near-optimal} shrinkage 
coefficient $\alpha^*$ for RTME.
In the addition, the ACVL method is amenable to parallel computation, and is directly applicable to sparse covariance
matrix estimation by means of thresholding the Regularized TME \cite{goes_lerman_nadler_2020}.
We demonstrate the efficiency and accuracy of the ACVL method on synthetic high-dimensional data sampled from
heavy-tailed elliptical distributions, as well as on real high-dimensional datasets for face recognition (Yale B), 
object recognition (CIFAR10 and CIFAR 100), and handwritten digit recognition (USPS).
Our experiments show that, with some additional mild computation, our proposed learning algorithm for shrinkage 
coefficient estimation is efficient and consistently more accurate than other methods in the literature.

An elementary proposal of our approach with some preliminary results appeared in \cite{karim_rtme_ieee_itw_2023}.
Our work here provides 
(\emph{i}) a detailed treatment for the theoretical motivation and derivation underlying the proposed approximation 
and algorithm, 
(\emph{ii}) a streamlined derivation for RTME for any desired target matrix, 
(\emph{iii}) a brief literature review for the different approaches for shrinkage coefficient estimation for RTME, and
(\emph{iv}) extensive experimental results on synthetic and real-world high-dimensional datasets.
The presentation of this work will proceed as follows.
Following the introduction, a concise review of different approaches for shrinkage coefficient estimation 
is discussed in Section (\ref{subsec:approaches_for_estimating_alpha}).
The notation used in this work, and the formal definition for elliptical distributions are introduced in 
Sections (\ref{subsec:notation_setup}) and (\ref{subsec:elliptical_distributions}), respectively.
Tyler's $M$-estimator (TME) and Regularized TME (RTME) are introduced in Section (\ref{sec:tme_rtme}).
The LOOCV approach for optimal shrinkage coefficient estimation is discussed in Section (\ref{sec:optimal_shrinkage_coeff}). 
In Section (\ref{sec:efficient_approx_acvl}) we present our proposed approximation for the LOOCV log-likelihood function. 
Empirical evaluations on simulated high-dimensional data from heavy-tailed elliptical distributions, and on 
real datasets in the context of face and object recognition are discussed in Section (\ref{sec:empirical_evaluation}). 
Concluding remarks and some future research directions are highlighted in Section (\ref{sec:conclusion}).


\subsection{Approaches for Shrinkage Coefficient Estimation for RTME}
\label{subsec:approaches_for_estimating_alpha}

As will be shown in the next section, the recursive nature of TME's estimating equation makes estimating 
an optimal shrinkage coefficient for RTME a non-trivial problem.
We note three different broad approaches were considered for shrinkage coefficient estimation for RTME: 
(\emph{i}) approaches based on oracle and \emph{random matrix theory} (RMT) results,
(\emph{ii}) approaches based on cross-validation techniques, and
(\emph{iii}) approaches based on the maximum likelihood principle.

Oracle-based approaches assume that the true scatter matrix $\mS$ is known and that the given samples 
are \emph{independent and identically distributed} (\emph{i.i.d}) realizations from a multivariate 
Gaussian distribution.
These methods proceed by defining an objective function that minimizes the mean squared error (MSE) between 
the true but unknown scatter matrix $\mS$ and the estimated regularized scatter matrix $\est{\mS}$.
Usually, these methods lead to closed-form solutions that are based on asymptotic estimates for the statistics needed 
for finding the optimal shrinkage coefficient \cite{tong_2018}.
Since the closed-form solution is a function of the unknown scatter matrix $\mS$, in practice, it is usually replaced
with the SCM, the trace normalized SCM, or a low-rank approximation of the SCM.
Oracle-based approaches were used in the works of 
Chen, Wiesel \& Hero \cite{chen_wiesel_hero_2011}, 
Ollila \& Tyler \cite{ollila_tyler_2014},  
Hoarau \etal \cite{hoarau_2017}, and 
Ashurbekova \etal \cite{ashurbekova_2021} for the general family of $M$-estimators.
Although oracle-based approaches are computationally efficient thanks to their closed-form solutions, they may come 
short in terms of accuracy due to the implicit assumptions in their asymptotic estimates and their reliance on the SCM.

Approaches based on RMT results are closely related to oracle-based methods.
In particular, RMT approaches are based on asymptotic analysis for regularized TME \emph{in the absence of
outliers}, and in the regime where both $n,p \goto \infty$ and $n/p \goto c$ for some constant $c \in (0,\infty)$.
RMT analysis for regularized TME was introduced by Couillet \& McKay \cite{couillet_mckay_2014} and studied for 
some problems in communications and finance 
\cite{kammoun_couillet_pascal_alouini_2018}. 
While RMT analysis for regularized TME provides insight into the asymptotic behavior of the estimator, RMT-based 
approaches are characterized by sophisticated computations that may not be efficient in practice 
(e.g. Proposition 2 in \cite{couillet_mckay_2014}) and may not yield unique solutions.
This has motivated Zhang \& Wiesel (ZW) \cite{zhang_wiesel_2016} to consider an alternative route to leverage 
the insights from RMT analysis.
In particular, based on the results in \cite{wiesel_geodesic_2012,zhang_cheng_singer_2016}, the Authors in 
\cite{zhang_wiesel_2016} modified the estimating equation for RTME -- and consequently its fixed point iterative 
algorithm -- to leverage the optimal and consistent estimator for the shrinkage coefficient developed by 
Ledoit \& Wolf in \cite{ledoit_wolf_2002,ledoit_wolf_2004}.
This makes ZW's approach more similar to oracle-based methods and its accuracy will be evaluated in 
\S \ref{sec:empirical_evaluation}.

Data-dependent approaches using CV techniques are based on:
(\emph{i}) choosing an appropriate loss function to be minimized with respect to $\alpha$, 
(\emph{ii}) grid search for the shrinkage coefficient, and 
(\emph{iii}) choosing one of the flavors of CV techniques which are computationally expensive but known to 
provide more accurate results in practice.
CV approaches were considered in the works of 
Abramovich \& Spencer \cite{abramovich_spencer_2007},
Wiesel \cite{wiesel_2012},
Sun, Babu \& Palomar \cite{sun_babu_palomar_2014}, and
Dumbgen \& Tyler \cite{dumbgen_tyler_2016}.
Shrinkage coefficient estimation using CV was also considered for the regularized SCM in the works of
Hoffbeck \& Landgrebe \cite{hoffbeck_landgrebe_96}, Theiller \cite{theiler_2012}, and Tong \etal \cite{tong_2018}.
These works have proposed fast algorithms for CV computations using efficient linear algebra-based
approximations.
Unfortunately, such efficient approximations cannot be directly leveraged in the context of RTME due the 
recursive nature of its estimating equations.

Finally, likelihood-based approaches are exemplified by the works of Besson \& Abramovich 
\cite{abramovich_besson_2_2013}. 
In \cite{abramovich_besson_2_2013}, the optimal shrinkage coefficient $\alpha$ is defined as the minimizer 
of a likelihood-ratio objective function that is parameterized by a low-rank scatter matrix; this low-rank 
scatter matrix is (itself) a function of the shrinkage coefficient $\alpha$.
The EL method was shown to be successful for some problems in wireless communications where it is reasonable 
to assume that the noisy samples have an elliptical distribution.
However, due to the specific well-controlled environments for such problems in wireless communications, 
the EL method may not be a generic approach for estimating an optimal shrinkage coefficient for problems settings 
in domains such as pattern recognition and computer vision. 


\subsection{Notation and Setup}
\label{subsec:notation_setup}

Scalars and indices are denoted by lowercase letters: $x, y$ and $i, j$, respectively.
Vectors are denoted by lowercase bold letters: $\vx, \vy$, and matrices by uppercase bold letters: $\mX, \mY$.
Sets are denoted by calligraphic letters: $\sX, \sY$, and spaces are denoted by double-bold uppercase letters:
$\mathbb{R}, \mathbb{S}$.
The identity matrix is denoted by $\matr{I}$, and $\vect{0}$ is the vector with all zeros, both with suitable 
dimensions from the context.
For $\vect{x} \in \RR^p$, $\norm{x}$ is the Euclidean norm.
For a matrix $\matr{A} = (a_{ij})$, $\norm{\matr{A}}_F$ is the Frobenius norm, $\trace{\matr{A}}$ is the matrix trace, 
and $\deter{\matr{A}}$ is the matrix determinant.
The space of symmetric and positive definite (PD) matrices is denoted by $\SS^p_+$.
The unit sphere in $\RR^p$ is denoted by $\mathcal{S}^p$, where 
$\mathcal{S}^p = \curbra{\vect{x} \in \RR^p ~\st ~\norm{\vect{x}} = 1}$.


\subsection{Elliptical Distributions}
\label{subsec:elliptical_distributions}

We will use the stochastic model due to Cambanis \etal \cite{cambanis_1981} and recently used in the literature to define \emph{elliptical} 
random vectors (RV) \cite{goes_lerman_nadler_2020}.
Let $\vz$ be a $p$ dimensional RV generated by the following model:
\begin{align}
  \vz 
  & = 
  \vmu + u\mS^{\mfrac{1}{2}} \vy
  =  
  \vmu + u\tilde{\vx} ~,
\label{eq:gen_model_elliptical_rv}
\end{align}
where $\vmu \in \RR^p$ is a location vector,
$\mS \in \SS^p_+$ is a \emph{scatter} or \emph{shape} matrix, 
$\vy$ is drawn uniformly from $\mathcal{S}^p$, and 
$u$ is a nonnegative random variable (r.v.) stochastically independent of $\vy$. 
The resulting RV $\vz$ from the model in (\ref{eq:gen_model_elliptical_rv}) is an \emph{Elliptically Distributed} (ED) RV.
Note that $\mS$ in (\ref{eq:gen_model_elliptical_rv}) is not unique since it can be arbitrarily scaled with $1/u$ absorbing the scaling factor $u$.
The distribution function of $u$, known as the \emph{generating distribution function}, constitutes the particular elliptical distribution family of the RV $\vz$.
If $\vz$ is an ED RV, its probability density function (PDF) is defined as:
\begin{align}
  f(\vz;\vmu,\mS,g_u) 
  & =
  \deter{\mS}^{-\mfrac{1}{2}}
  g_u\bra{\bar{\vz}^\top \mS^{-1} \bar{\vz}},
  \label{eq:elliptical_pdf}
\end{align}
where $\bar{\vz} = (\vz - \vmu)$, and $g_u:\RR_+ \mapsto \RR_+$ is a nonnegative decreasing function known as the 
\emph{density generator function} and is not dependent on $\vmu$ and $\mS$, but dependent on the generating
distribution function of $u$.
The density generator function determines the shape of the PDF, as well as the \emph{tail decay} of the distribution.
For any elliptical distribution, if its population covariance matrix $\MatrSigma$ exists, then $\MatrSigma = c_g\mS$ 
for some constant $c_g > 0$ that is dependent on $g_u$.


\section{Regularized Tyler's $M$-Estimator (RTME)}
\label{sec:tme_rtme}

Let $\sZ_n = \bra{\vz_1,\dots,\vz_n}$ be a sample of $n$ \emph{independent} and \emph{identically distributed} (\iid) realizations 
from the model in (\ref{eq:gen_model_elliptical_rv}) with location vector $\vmu = \vect{0}$ and scatter matrix $\mS$.
We are interested in computationally efficient and statistically accurate algorithms for estimating the population scatter matrix 
$\mS$ using the samples in $\sZ_n$ in the setting where $p > n$.
Here we do not make \apriori sparsity assumptions on the scatter matrix $\mS$.
Without any \apriori knowledge on $c_g$ and $g_u$, it may seem less probable to obtain a good estimator for $\mS$. 
In addition, for some elliptical distributions -- such as the multivariate Cauchy distribution -- they may have infinite second 
moments in which case the population covariance matrix $\MatrSigma$ does not exist.
Thus, it may always be better to consider and estimate the \emph{normalized} scatter matrix $\mS$ which is always defined \cite{pascal_chitour_queck_2014}.

TME can be derived as an ML estimator of the shape matrix for the \emph{Angular Central Gaussian} (ACG) distribution (defined in Equation \ref{eq:acg_density_model})
based on the sample $\sZ_n$ \cite{tyler_acg_tme_1987}.
With $\vmu = \vect{0}$, the sample $\sZ_n$ can be written as $\bra{u_1\tilde{\vx}_1,\dots,u_n\tilde{\vx}_n}$.
Since the scalars $u_1,\dots,u_n$ are unknown, there is a scaling ambiguity and one can only expect to estimate 
matrix $\mS$ up to a scaling factor.
TME overcomes this limitation by working with the normalized samples:
$\vx_i = \vz_i/\norm{\vz_i} = \tilde{\vx}_i/\norm{\tilde{\vx}_i}$, $1 \leq i \leq n$, where the scalars $u_i$ 
cancels out.
The PDF for the vectors $\vx_1,\dots,\vx_n$ is given by:
\begin{align}
  f\bra{\vx;\mS}
  & = 
  (2\pi)^{-\mfrac{p}{2}}\Gamma(\sfrac{1}{2})
  \det\bra{\mS}^{-\mfrac{1}{2}}
  \bra{\vect{x}^\top \mS^{-1} \vect{x}}^{-\mfrac{p}{2}},
   \label{eq:acg_density_model}
\end{align}
where $\vx \in \sS^p$, $\Gamma(\cdot)$ is the Gamma function, and $\Gamma(p/2) / (2\pi)^{\frac{p}{2}}$ is the surface area of $\sS^p$.
The ACG density in (\ref{eq:acg_density_model}) represents the \emph{distribution of directions} for samples drawn from a multivariate 
Gaussian distribution with zero mean and covariance matrix $\mS$ \cite{tyler_acg_tme_1987}. 
Thus, only the directions of outliers can affect TME's performance but not their magnitude.
Given an \iid random sample $\sX_n = \bra{\vx_1,\dots,\vx_n}$ from a distribution having the ACG density in 
(\ref{eq:acg_density_model}), the likelihood of $\sX_n$ with respect to $\mS$ is \emph{proportional} to:
\begin{align}
  \lhood{\sX_n;\mS}
  & = 
  \deter{\mS}^{-n/2}
  \prod_{i=1}^n
  \bra{\vect{x}_i^\top \mS^{-1} \vect{x}_i}^{-\frac{p}{2}} ~.
  \label{eq:ml_of_s}
\end{align}
Taking $-\log$ of $\lhood{\sX_n;\mS}$ yields the following loss function which will be needed for our following discussions:
\begin{align}
    \loglhood{\sX_n;\mS}
    & = 
    \frac{p}{2}\sum_{i=1}^n \log\bra{\vect{x}_i^\top \mS^{-1} \vect{x}_i}
    +
    \frac{n}{2}\log\deter{\mS}.
    \label{eq:loglhood_data_sctermatrix}
\end{align}
Taking the derivative of $\loglhood{\sX_n;\mS}$ with respect to $\mS$ and equating it to zero, the ML estimator 
for $\mS$ is the solution to the following fixed point equation:
\begin{align}
  \mS_n 
  & =
  \frac{p}{n}
  \sum_{i=1}^n
  \vx_i\vx_i^\top/(\vx_i^\top \mS_n^{-1} \vx_i) ~,
  \label{eq:tme_fixed_point_eq}
\end{align}
where $\vx_i \neq \vect{0}$ for $i = 1,\dots,n$ since samples lying at the origin provide no directional information on the scatter matrix.
If $n > p(p-1)$, Theorem 1 in \cite{tyler_acg_tme_1987} states that with probability one, the ML estimate of $\mS$ \emph{exists}, corresponds 
to the solution in (\ref{eq:tme_fixed_point_eq}), and is \emph{unique} up to a positive multiplicative scalar. 
The solution to (\ref{eq:tme_fixed_point_eq}) can be found using the following fixed point iteration (FPI) algorithm:
\begin{align}
  \est{\mS}_{t+1} 
  & =
  \frac{p}{n}
  \sum_{i=1}^n
  \vx_i\vx_i^\top / (\vx_i^\top \est{\mS}_{t}^{-1} \vx_i) ~,
  \label{eq:tme_fixed_point_algo}
\end{align}
with $\est{\mS}_0 = \mI$, or any arbitrary initial $\est{\mS}_0 \in \SS^p_+$ \cite{kent_tyler_1988}.
Theorem 2.2 and Corollaries 2.2 \& 2.3 in \cite{tyler_tme_1987} show that if $n > p+1$ and assuming that every $p$ samples out of $\sX_n$ 
are \emph{linearly independent} with probability one, and that the maximum likelihood estimate of $\mS$ exists, then the FPI algorithm in 
(\ref{eq:tme_fixed_point_algo}) \emph{almost surely} converges to the solution of (\ref{eq:tme_fixed_point_eq}), and the limiting solution 
$\est{\mS} = \est{\mS}_T$ computed at the last iterate $T$ is unique up to a positive multiplicative scalar. 

TME has various attractive properties and is asymptotically optimal under different criteria.
In particular, for elliptically distributed data, TME is the \emph{most robust} estimator for the scatter matrix in a \emph{minimax} sense; 
minimizing the maximum asymptotic variance (see Remark 3.1 in \cite{tyler_tme_1987}).
Further, for elliptical distributions, Theorem 3.3 in \cite{tyler_tme_1987} states that the asymptotic distribution of $\mS_n$ does not 
depend on the specific form of the density generator function $g_u$ in (\ref{eq:elliptical_pdf}); i.e. it is \emph{distribution-free} within 
the class of elliptical distributions. 
Last, strong consistency and asymptotic normality for TME are established in Theorems 3.1 \& 3.2 in \cite{tyler_tme_1987}, respectively.

Unfortunately, when $p > n$, TME is not defined; the LHS of (\ref{eq:tme_fixed_point_eq}) must be a full rank symmetric PD matrix, while the 
RHS is rank-deficient.\footnote{For TME, regularization may still be needed for $p \leq n \leq p(p-1)$ when the points are not in general 
position, and/or the samples are not drawn from an elliptical distribution.}
Various researchers have proposed different flavors of RTME using the spirit of Ledoit \& Wolf \cite{ledoit_wolf_2004} 
linear shrinkage estimator \cite{abramovich_spencer_2007,chen_wiesel_hero_2011,wiesel_2012,pascal_chitour_queck_2014,ollila_tyler_2014}.
In particular, Sun, Babu \& Palumar (SBP) \cite{sun_babu_palomar_2014} proposed the following penalized log-likelihood function to derive 
a regularized version of TME:
\begin{equation}
\begin{aligned}
  \sL_{\mathcal{P}}(\sX_n;\mS)
  & = 
  \loglhood{\sX_n;\mS} + \beta\mathcal{P}(\mS) ~,
\end{aligned}
\label{eq:pen_logll}
\end{equation}
where $\sL_{\mathcal{P}}(\sX_n;\mS)$ is defined in (\ref{eq:loglhood_data_sctermatrix}), and $\mathcal{P}(\mS)$ is the 
penalty function defined as:
\begin{align}
  \mathcal{P}(\mS)
  & =
  \trace{\mS^{-1}\mT}
  +
  \log\deter{\mS} ~,
\end{align}
with $\beta > 0$ is the regularization parameter (or shrinkage coefficient) and $\mT \in \SS_p^+$ is a given 
target matrix that has some desirable structural properties (diagonal, banded, Toeplitz, etc.).
Letting $\alpha = \beta/(1 + \beta)$, the solution to (\ref{eq:pen_logll}) has to satisfy the fixed point equation:
\begin{align}
  \mS_n
  & =
  (1 - \alpha)
  \frac{p}{n}
  \sum_{i = 1}^n 
  \frac{\vx_i \vx_i^\top}{\vx_i^\top \mS_n^{-1} \vx_i}
  + 
  \alpha
  \mT ~.
  \label{eq:rtme_fixed_point_eq}
\end{align}
Note that $\alpha \in (0,1)$ for any $0 < \beta < \infty$.
Starting from an arbitrary $\est{\mS}_0 \in \SS_p^+$, the final solution can be obtained using the following 
\emph{Regularized} FPI (RFPI) algorithm: 
\begin{align}
  \est{\mS}_{t+1}(\alpha)
  & =
  (1 - \alpha)
  \frac{p}{n}
  \sum_{i = 1}^n 
  \frac{\vx_i \vx_i^\top}{\vx_i^\top \est{\mS}_t^{-1}(\alpha) \vx_i}
  + 
  \alpha
  \mT ~,
  \label{eq:rtme_fixed_point_algo}
\end{align}
where $\alpha \in (0, 1)$ is the \emph{shrinkage coefficient} that controls the amount of shrinkage applied 
to scatter matrix $\mS$ towards the target matrix $\mT$.
Theorem 11 and Proposition 13 in \cite{sun_babu_palomar_2014} establish the necessary and sufficient conditions 
for the \emph{existence} and \emph{uniqueness} of the solution to Equation (\ref{eq:rtme_fixed_point_eq}), while
Proposition 18 ensures that the RFPI algorithm in (\ref{eq:rtme_fixed_point_algo}) converges to the unique solution 
of (\ref{eq:rtme_fixed_point_eq}).

Without loss of generality, if $\mT = \mI$, $\alpha = 0$, one restores the unbiased TME in (\ref{eq:tme_fixed_point_algo}), 
and if $\alpha = 1$ the estimator reduces to the uncorrelated scatter matrix $\alpha\mI$.
When $p < n$, and the samples are drawn from an elliptical distribution, $\alpha$ is expected to be zero 
(or close to zero) and results for the existence and uniqueness of the estimator still hold \cite{pascal_chitour_queck_2014}.
When $p \geq n$, $\alpha$ is expected to be large; however to ensure the \emph{existence and uniqueness} of the
estimator, $\alpha$ needs to be strictly greater than $1 - n/p$ \cite{pascal_chitour_queck_2014,sun_babu_palomar_2014}.


\subsection{Runtime Analysis}
\label{subsec:computational_complexity}

The magnitude of $\alpha$ has an impact on the accuracy of the final estimate $\est{\mS} = \est{\mS}_T$, as well as 
on the convergence speed for the RFPI algorithm.
In particular, for $p \geq n$, Lemma 1 in \cite{goes_lerman_nadler_2020} gives a result on the \emph{uniform linear convergence} 
of the algorithm in (\ref{eq:rtme_fixed_point_algo}) to a unique solution; for desired accuracy $\gvareps$, 
convergence ratio $r$, and sufficiently large $\alpha > 1 - n/p$, \emph{at most} $\ceil{\log_{1/r}(1/\gvareps)}$ 
iterations are needed for (\ref{eq:rtme_fixed_point_algo}) to converge to the unique solution of 
(\ref{eq:rtme_fixed_point_eq}).

A preliminary analysis of the RFPI algorithm shows that the running time for each iteration is $O(np^2 + p^3)$ 
where $O(np^2)$ is the time needed to compute the sum of rank-one matrices, and $O(p^3)$ is the time needed to compute 
the inverse matrix $\est{\mS}^{-1}_{t}(\alpha)$.
Since $\est{\mS}_{t}(\alpha)$ is PD, an efficient computation for the inverse can be done using Cholesky factorization 
\cite{book_matrix_comp_golub_1996}: 
$\est{\mS}_{t}(\alpha) = \mL\mL^\top$, where $\mL$ is a lower triangular matrix. 
Cholesky factorization requires $\sfrac{1}{3}p^3$ flops: $\sfrac{1}{6}p^3$ multiplications, and $\sfrac{1}{6}p^3$ 
additions. 
Finally, inverting a triangular matrix will require $p^2$ flops. 
If $T$ iterations are needed for the RFPI algorithm to converge, its total running time complexity will be 
$O(T(np^2 + p^3))$.
If $n \gg p$, then RFPI's runtime complexity is dominated by the sum of rank-one matrices; i.e. $O(np^2T)$. 
While if $p \gg n$, then RFPI's runtime complexity is dominated by the matrix inversion step; i.e. $O(p^3T)$.


\section{Optimal Choice of Shrinkage Coefficient $\alpha$}
\label{sec:optimal_shrinkage_coeff}

Our objective is to find an appropriate $\alpha$ that is \emph{optimal} under a suitable loss function.
In particular, if $p \ll n$ and the samples are drawn from an elliptical distribution, we expect $\alpha$ to be zero 
(or close to zero). On the contrary, if $p \geq n$, we expect that a larger $\alpha$ will be more suitable in this case.
Even when $p < n$ and the samples are heavy-tailed and not from an elliptical distribution, it is expected that 
$\alpha$ will be large. 
If the true scatter matrix $\mS$ is known, one can choose a shrinkage coefficient that minimizes an appropriate 
distance metric between $\est{\mS}$ and $\mS$.
Since $\mS$ is unknown, our approach will depend on the likelihood function of $\sX_n$ with respect to $\mS$ in 
(\ref{eq:ml_of_s}).
In particular, for a \emph{fixed} $\bar{\alpha} \in (0,1)$, suppose that $\est{\mS}(\bar{\alpha})$ is an estimate of 
the true scatter matrix $\mS$.
Given the sample $\sX_n$, one can assess the quality of $\est{\mS}(\bar{\alpha})$ with respect to $\sX_n$ using the 
likelihood function $\lhood{\sX_n;\mS}$ in (\ref{eq:ml_of_s}); or equivalently using the loss function $\loglhood
{\sX_n;\mS}$ in (\ref{eq:loglhood_data_sctermatrix}), by replacing $\mS$ with $\est{\mS}(\bar{\alpha})$.
Using this approach, an optimal $\alpha$ with respect to $\sX_n$, denoted $\alpha^*$, will be the one that minimizes 
$\sL(\sX_n,\est{\mS}(\alpha))$ over $\alpha \in (0,1)$; i.e.
\begin{align}
    \alpha^*
    & = 
    \argmin{\alpha \in (0,1)}
    \quad
    \sL(\sX_n ; \est{\mS}(\alpha)) ~.
    \label{eq:opt_scatter_rtme_nll}
\end{align}
The problem with this direct approach is that $\est{\mS}(\alpha)$ needs to be computed using the sample $\sX_n$.
That is, the sample $\sX_n$ will be used twice; first time to compute $\est{\mS}(\alpha)$, and a second time to assess 
the quality of $\est{\mS}(\alpha)$ using $\sL(\sX_n ; \est{\mS}(\alpha))$ in (\ref{eq:loglhood_data_sctermatrix}).
This is known as \emph{double-dipping}, and inevitably it leads to an \emph{overfit} estimate of the shrinkage 
coefficient $\alpha$.

CV techniques overcome this problem by splitting the data into two non-overlapping samples; 
one sample for estimating $\mS$ and the other sample for estimating the loss $\sL$.
Here, we propose to use \emph{Leave-One-Out} CV (LOOCV) for estimating $\mS$ and $\sL$ 
\cite{cover_1969}.
In particular, for $1 \leq i \leq n$, LOOCV splits $\sX_n$ into two sub-samples: the sample 
$\sX_{n \setminus i} = (\vx_1,\dots,\vx_{i-1},\vx_{i+1},\dots,\vx_n)$, and the sample $(\vx_i)$ which contains 
the single data point $\vx_i$.
The sample $\sX_{n \setminus i}$ will be used to estimate $\mS(\alpha)$ using the RFPI algorithm in 
(\ref{eq:rtme_fixed_point_algo}), while $(\vx_i)$ will be used to estimate $\sL(\vx_i ; \est{\mS}(\alpha))$.
This process is repeated $n$ times and the LOOCV estimate will be the average of all $\sL(\vx_i ; \est{\mS}(\alpha))$, 
$1 \leq i \leq n$.
Using LOOCV, an optimal $\alpha$ can be computed as follows:
\begin{align}
    \est{\alpha}^*_\CV
    & =
    \argmin{\alpha \in (0,1)}
    \quad 
    \sL_{\CV}(\sX_n,\alpha) ~, 
    \label{eq:opt_alpha_exact_loocv}
\end{align}
where $\sL_{\CV}(\cdot)$ is the average CV \emph{Loss} (CVL) defined as:
\begin{align}
    \sL_{\CV}(\sX_n,\alpha)
    & = 
    \frac{1}{n}
    \sum_{i = 1}^n
    \sL(\vx_i ; \est{\mS}(\alpha ; \sX_{n \setminus i})) ~,
    \label{eq:aloocvnll}
\end{align}
and $\est{\mS}(\alpha ; \sX_{n \setminus i})$ is estimated from the points in $\sX_{n \setminus i}$ using 
the RFPI algorithm (\ref{eq:rtme_fixed_point_algo}). 
In practice, one possible approach to solve problem (\ref{eq:opt_alpha_exact_loocv}) can be using a simple
\emph{grid search}: 
(\emph{i}) define a discrete range of increasing values of $\alpha$: $(\alpha_1,\dots,\alpha_j,\dots,\alpha_m)$; 
(\emph{ii}) evaluate $\sL_{\CV}(\sX_n,\alpha_j)$ for each $\alpha_j$ using (\ref{eq:aloocvnll}); and 
(\emph{iii}) choose $\alpha_j$ with the minimum $\sL_{\CV}(\cdot)$.\footnote{Note that when $p > n$, and for existence 
and uniqueness results to hold, $\alpha$ needs to be strictly greater than $1 - n/p$ 
\cite{pascal_chitour_queck_2014,sun_babu_palomar_2014}, and hence there is no need to evaluate $\sL_{\CV}(\cdot)$ for
$\alpha \leq 1 - n/p$.}
For a \emph{reasonably fine} discretization for the range of $\alpha$'s, this direct estimation approach 
will yield an estimate for $\alpha$ that is \emph{reasonably close} to its optimal value.
With little abuse of terminology, and for reasons that will be discussed shortly, we refer to this method for 
estimating $\alpha^*$ as the \emph{Exact CVL} method. 


\subsection{Properties of LOOCV and its Computational Overhead}
\label{subsec:prop_loocv}

The Riemannian manifold of symmetric PD matrices $\SS^p_+$ is a subset of $\RR^{p(p+1)/2}$ and is a compact 
space \cite{kent_tyler_1988}.
The log likelihood function $\sL(\sX_{n} ; \mS)$ in (\ref{eq:loglhood_data_sctermatrix}) is geodesically convex with 
respect to $\SS^p_+$ \cite{wiesel_geodesic_2012,dumbgen_tyler_2016}, and properties for this type of likelihood 
functions has been studied in \cite{book_amari_nagaoka_00}.
In particular, $\sL(\sX_{n} ; \mS)$ maintains the three main properties of maximum likelihood 
estimators: \emph{consistency}, \emph{efficiency}, and \emph{functional invariance}.
On the other hand, the LOOCV estimate is \emph{almost an unbiased} estimate in the following sense \cite[Ch. 24]{book_devroye_gyorfi_lugosi_1996}: 
for fixed $p$ and $\bar{\alpha}$,
\begin{align}
  \EE \sL_{\CV}(\sX_n,\bar{\alpha}) &= \EE \sL(\sX_{n-1}^{'} ; \est{\mS}(\bar{\alpha} ; \sX_{n-1})) ~,
  \nonumber  
\end{align}
where the expectations are w.r.t the random samples $\sX_n$, $\sX_{n-1}^{'}$, $\sX_{n-1}$, and 
$\sX_{n-1}^{'} \indep \sX_{n-1}$. That is, $\sL_{\CV}(\sX_n,\bar{\alpha})$ is an estimator for 
$\sL^{*}(\est{\mS}(\bar{\alpha} ; \sX_{n-1}))$ 
rather than for 
$\sL^{*}(\est{\mS}(\bar{\alpha} ; \sX_{n}))$,
where 
\begin{align}
  \sL^{*}(\est{\mS}(\bar{\alpha} ; \sX_{n-1}))
  & = 
  \EE[\sL(\sX_{n-1}^{'} ; \est{\mS}(\bar{\alpha} ; \sX_{n-1})) ~|~ \sX_{n-1}] ~,~ \mathrm{and}
  \nonumber
  \\
  \sL^{*}(\est{\mS}(\bar{\alpha} ; \sX_n))
  & = 
  \EE[\sL(\sX_n^{'} ; \est{\mS}(\bar{\alpha} ; \sX_n)) ~|~ \sX_n] .
  \nonumber
\end{align} 
The random quantities 
$\sL^{*}(\est{\mS}(\bar{\alpha} ; \sX_{n-1}))$ 
and
$\sL^{*}(\est{\mS}(\bar{\alpha} ; \sX_n))$
converge with probability one, and thus for large values of $n$ the difference between
$\sL^{*}(\est{\mS}(\bar{\alpha} ; \sX_n))$
and 
$\sL^{*}(\est{\mS}(\bar{\alpha} ; \sX_{n-1}))$
will be negligible.
The asymptotic properties of $\sL(\sX_{n} ; \mS)$ encourage us to postulate the following proposition which will be 
useful for introducing our approximation approach discussed in \S \ref{sec:efficient_approx_acvl}.

\begin{proposition}
\label{prop:main}
Let $\vx = \frac{\vz}{\norm{\vz}}$ be a random vector from the model in (\ref{eq:gen_model_elliptical_rv}) s.t.
$\vx$ is independent of $\sX_n$, and let $p$ and $\bar{\alpha}$ be predefined fixed values.
Then, under the \iid assumption for the samples in $\sX_n$ and from the consistency of $\sL(\sX_{n},\mS)$,  
we have that for large values of $n$, the difference between 
$\sL(\vx ; \est{\mS}(\bar{\alpha} ; \sX_n))$ 
and 
$\sL(\vx ; \est{\mS}(\bar{\alpha} ; \sX_{n \setminus i}))$
will be small for any $i$ chosen (randomly) from $i = 1,\dots,n$.
\end{proposition}

Proposition (\ref{prop:main}) postulates, based on an asymptotic argument, that for a fixed $p$ and $\bar{\alpha}$, 
and as $n$ is increasing, the difference between 
$\sL(\vx ; \est{\mS}(\bar{\alpha} ; \sX_n))$
and
$\sL(\vx ; \est{\mS}(\bar{\alpha} ; \sX_{n \setminus i}))$
will be small for any sample $\vx_i$ randomly chosen from $\sX_n$, where $i = 1,\dots, n$.
In particular, Proposition (\ref{prop:main}) implies that under the \iid assumption for $\sX_n$, and for large $n$,
$\est{\mS}(\bar{\alpha} ; \sX_n) \approx \est{\mS}(\bar{\alpha} ; \sX_{n \setminus i})$, or more generally,
$\est{\mS}(\bar{\alpha} ; \sX_n) \approx \est{\mS}(\bar{\alpha} ; \sX_{n-1})$; 
i.e. the estimator for $\mS$ is \emph{not too sensitive} to the removal of one sample from $\sX_n$.
The notion of sensitivity of an estimator with respect to the removal (or replacement) of one sample from $\sX_n$
is known as \emph{algorithmic stability} and it has been extensively leveraged in learning theory to derive 
generalization bounds on the risk of various learning algorithms 
\cite{kearns_ron_neco_99,karim_csaba_alt19}.
Our approximation approach introduced in the following section will leverage the previous insight from Proposition 
(\ref{prop:main}) to approximate $\est{\mS}(\bar{\alpha} ; \sX_{n \setminus i})$ and speedup the computation for the 
LOOCV estimate in $(\ref{eq:aloocvnll})$.

LOOCV is notorious for its high computational overhead.
Indeed, for a fixed $\bar{\alpha}$ and for $n$ samples in $\sX_n$, LOOCV will make $n$ calls for the RFPI 
algorithm in order to compute $\sL(\vx_i,\est{\mS}(\alpha ; \sX_{n \setminus i}))$ in the RHS of (\ref{eq:aloocvnll}).
Thus, for $m$ values of $\alpha_j$ from $(\alpha_1,\dots,\alpha_m)$, the Exact CVL method in 
(\ref{eq:opt_alpha_exact_loocv}) will require $mn$ calls for the RFPI algorithm, which is prohibitive even 
for moderate values of $n$.
If the RFPI algorithm requires $T$ iterations to converge, it will consume $O(mn * T(np^2 + p^3))$ time from the Exact 
CVL method in (\ref{eq:opt_alpha_exact_loocv}), where $O(T(np^2 + p^3))$ is the running time for a single call for the 
RFPI algorithm.

\begin{figure*}[t]
\makebox[\textwidth][c]{\includegraphics[trim=7 3 5 2,clip,width=\textwidth]{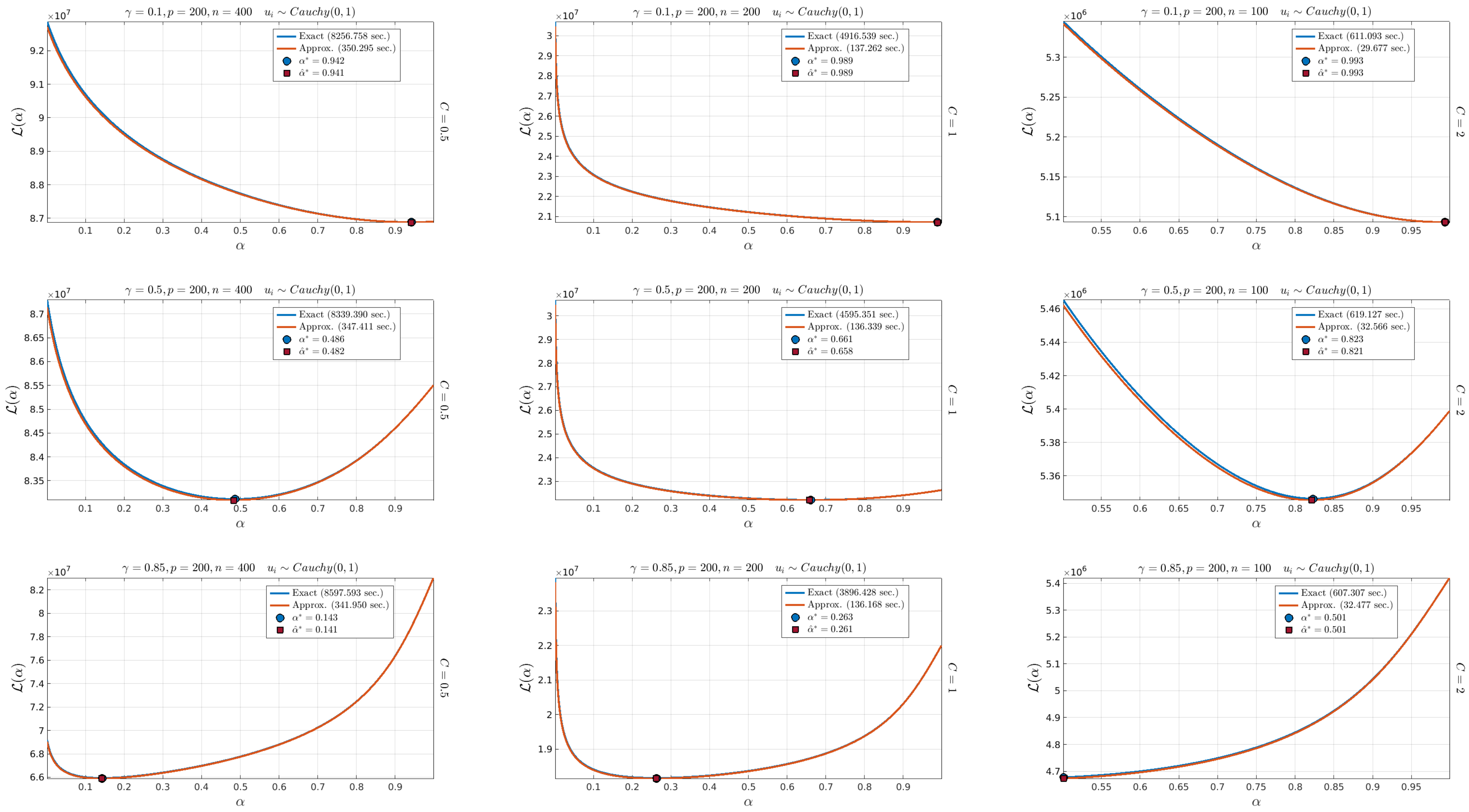}}
\caption{Comparison between \emph{Exact} and \emph{Approximate} CVL for samples drawn from a multivariate Cauchy
distribution in three different settings; $p < n$ (left), $p = n$ (middle), and $p > n$ (right), and for three 
different values of $\gamma=\{0.1, 0.5, 0.85\}$. The blue circle and red square indicate the optimal values for 
$\alpha$ obtained from the Exact and Approximate CVL methods, respectively. 
The running times (in seconds) for the Exact and Approximate CVL methods are shown in the legend.
Speedup for the Approximate CVL method over the Exact CVL method for each sub-figure is shown in Table \ref{tab:speedup_synthetic_data} .} 
\label{fig:exp07_optalpha_exact_vs_aloocv_cauchy}
\end{figure*}

\begin{figure*}[t]
\makebox[\textwidth][c]{\includegraphics[trim=7 3 5 2,clip,width=\textwidth]{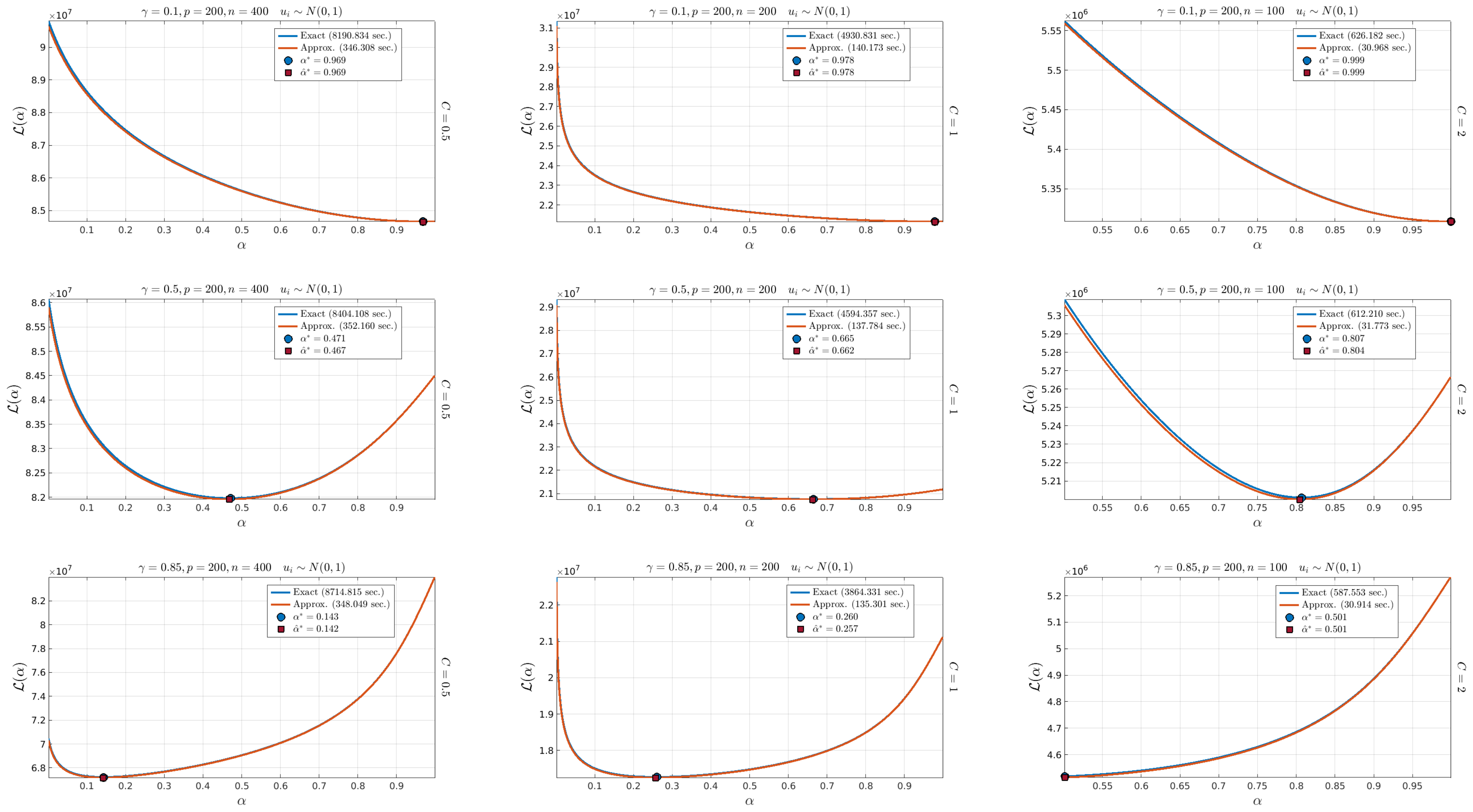}}
\caption{Comparison between \emph{Exact} and \emph{Approximate} CVL for samples drawn from a multivariate Gaussian 
distribution in three different settings; $p < n$ (left), $p = n$ (middle), and $p > n$ (right), and for three 
different values of $\gamma=\{0.1, 0.5, 0.85\}$. The blue circle and red square indicate the optimal values for 
$\alpha$ obtained from the Exact and Approximate CVL methods, respectively.
The running times (in seconds) for the Exact and Approximate CVL methods are shown in the legend.
Speedup for the Approximate CVL method over the Exact CVL method for each sub-figure is shown in Table \ref{tab:speedup_synthetic_data} .} 
\label{fig:exp07_optalpha_exact_vs_aloocv_gaussian}
\end{figure*}

Our objective in the following section is to reduce the time consumed by the RFPI algorithm in the Exact CVL method 
by a factor of $n$; to be $O(m * T(np^2 + p^3))$ instead of $O(mn * T(np^2 + p^3))$. 
In particular, we propose an efficient approximation for $\est{\mS}(\alpha,\sX_{n \setminus i})$ in 
(\ref{eq:aloocvnll}) so that the RFPI algorithm is invoked $m$ times only instead of $mn$ times to compute 
$\sL_{\CV}(\sX_n,\alpha)$ in (\ref{eq:opt_alpha_exact_loocv}).
The gain in speed due to this approximation while maintaining the accuracy of the estimated $\alpha$ is depicted
in Figures (\ref{fig:exp07_optalpha_exact_vs_aloocv_cauchy}) and (\ref{fig:exp07_optalpha_exact_vs_aloocv_gaussian})
for two elliptical distributions, the multivariate Cauchy distribution, and the multivariate Gaussian distribution,
respectively.
In particular, Figures (\ref{fig:exp07_optalpha_exact_vs_aloocv_cauchy}) \& (\ref{fig:exp07_optalpha_exact_vs_aloocv_gaussian})
compare the \emph{Exact} CVL method with the \emph{approximation} developed in the following section in terms of the average CV 
loss in (\ref{eq:aloocvnll}), running time (in seconds), and the optimal $\alpha$ obtained from each method (more details in 
\S\ref{sec:empirical_evaluation}).

\section{Efficient Approximation of $\est{\mS}(\alpha,\sX_{n \setminus i})$}
\label{sec:efficient_approx_acvl}

The approximation approach proposed here is motivated by our the observation from Proposition (\ref{prop:main}) that
under the $\iid$ assumption on $\sX_n$, and for large $n$, the estimator for $\mS$ is not too sensitive to the removal
of one sample from $\sX_n$; i.e. 
$\est{\mS}(\bar{\alpha} ; \sX_n) \approx \est{\mS}(\bar{\alpha} ; \sX_{n-1})$.
For a fixed $\bar{\alpha}$, the RFPI algorithm in (\ref{eq:rtme_fixed_point_algo}) can be expressed as follows:
\begin{align}   
  \est{\mS}_{t+1}(\bar{\alpha})
  & = 
  (1 - \bar{\alpha})
  p
  \bra{ 
  \frac{1}{n}
  \sum_{i=1}^n
  w_{t,i}^{-1} \vx_i\vx_i^\top} 
  +
  \bar{\alpha}\mT
  ~,
  \label{eq:rtme_fixed_point_algo2}
\end{align}
where $w_{t,i} = \vx_i^\top \est{\mS}_{t}^{-1}(\bar{\alpha}) \vx_i$, and  $t=1,\dots,T$.
That is, the first term in the RHS of (\ref{eq:rtme_fixed_point_algo2}) involves a weighted 
sample covariance matrix using the weights $w_{t,i}$ and the RFPI algorithm iteratively 
estimates these weights until convergence.
For initial matrix $\est{\mS}_0 \in \SS^p_+$, let $(\est{w}_1,\est{w}_2,\dots,\est{w}_n)$ 
be the optimal weights estimated using $\sX_n$ and the RFPI in (\ref{eq:rtme_fixed_point_algo2}).
Then, the \emph{final} estimate for the scatter matrix can be written as:
\begin{align}
  \est{\mS}(\bar{\alpha};\sX_n)
  & = 
  (1 - \bar{\alpha})
  \frac{p}{n}
  \sum_{i=1}^n
  \frac{1}{\est{w}_i}\vx_i\vx_i^\top 
  +
  \bar{\alpha}\mT
  ~.
  \label{eq:final_rtme_bign}
\end{align}
Let $\sX_{n \setminus i} = (\vx_1,\dots,\vx_{i-1},\vx_{i+1},\dots,\vx_n)$.
Similar to (\ref{eq:final_rtme_bign}), using $\bar{\alpha}$ and initial matrix $\est{\mS}_0$, the \emph{final} 
estimate for the scatter matrix using $\sX_{n \setminus i}$ and the RFPI in (\ref{eq:rtme_fixed_point_algo2}) 
will be:
\begin{align}
  \est{\mS}(\bar{\alpha};\sX_{n \setminus i})
  & = 
  (1 - \bar{\alpha})
  \frac{p}{n-1}
  \sum_{\underset{j \neq i}{j = 1}}^{n}
  \frac{1}{\est{v}_j}\vx_j\vx_j^\top 
  +
  \bar{\alpha}\mT
  ~,
  \label{eq:final_rtme_smalln}
\end{align}
where 
$(\est{v}_{1},\dots,\est{v}_{i-1},\est{v}_{i+1},\dots,\est{v}_{n})$ are the optimal weights estimated using 
$\sX_{n \setminus i}$.
From Proposition (\ref{prop:main}), and using initial $\est{\mS}_0$ to obtain the \emph{final} estimates 
in (\ref{eq:final_rtme_bign}) and (\ref{eq:final_rtme_smalln}), it is expected that for large $n$:
$\est{\mS}(\bar{\alpha} ; \sX_n) \approx \est{\mS}(\bar{\alpha} ; \sX_{n \setminus i})$, and the difference
between 
$\sL(\vx ; \est{\mS}(\bar{\alpha} ; \sX_n))$ 
and 
$\sL(\vx ; \est{\mS}(\bar{\alpha} ; \sX_{n \setminus i}))$
will be a arbitrarily small.
In terms of computations, and for a fixed $\bar{\alpha} \in (0,1)$, computing the final estimate 
$\est{\mS}(\bar{\alpha};\sX_{n \setminus i})$ for each $i = 1,\dots,n$ requires invoking the RFPI algorithm 
$n$ times during the LOOCV procedure. 
This yields a total running time of $O(nT(np^2 + p^3))$ which is inefficient even for moderate values of 
$n$ and $p$.

To introduce our proposed approximation, suppose that the true scatter matrix $\mS^* \in \SS^+_p$ is known and 
$(\mS^*)^{-1}$ has been computed.
Then, the final estimate $\est{\mS}(\bar{\alpha};\sX_n)$ in (\ref{eq:final_rtme_bign}) can be 
\emph{directly} computed without invoking the RFPI algorithm in (\ref{eq:rtme_fixed_point_algo2}):
\begin{align}
  \est{\mS}(\bar{\alpha};\sX_n)
  & = 
  \frac{(1 - \bar{\alpha})p}{n}
  \sum_{i=1}^n
  \frac{1}{\est{w}^*_i}\vx_i\vx_i^\top 
  +
  \bar{\alpha}\mT
  ,\,\textrm{\small where }
  \label{eq:final_rtme_n_true_s} 
  \\
  \est{w}^*_i 
  & = 
  \vx_i^\top (\mS^*)^{-1} \vx_i 
  ~.
  \nonumber
\end{align}
Similarly, the final estimate $\est{\mS}(\bar{\alpha};\sX_{n \setminus i})$ in 
(\ref{eq:final_rtme_smalln}) can be \emph{directly} computed without invoking the RFPI algorithm in 
(\ref{eq:rtme_fixed_point_algo2}):
\begin{align}
\est{\mS}(\bar{\alpha};\sX_{n \setminus i})
& = 
\frac{(1 - \bar{\alpha})p}{n - 1}
\sum_{\underset{j \neq i}{j = 1}}^n
\frac{1}{\est{v}^*_j}\vx_j\vx_j^\top 
+
\bar{\alpha}\mT
,\, \textrm{\small where }
\label{eq:final_rtme_nm1_true_s}  
\\
\est{v}^*_j
& = 
\vx_j^\top (\mS^*)^{-1} \vx_j 
~.
\nonumber
\end{align}
Note that both 
$\est{w}_i^*$ in (\ref{eq:final_rtme_n_true_s}) and 
$\est{v}_j^*$ in (\ref{eq:final_rtme_nm1_true_s}) 
are dependent on the true but unknown scatter matrix $\mS^*$ and in this case: 
$\est{v}_j^* = \est{w}^*_j$ for $j \neq i$, and $j = 1,\dots,n$.
Since $\mS^*$ is unknown, we propose to approximate 
$\est{\mS}(\bar{\alpha};\sX_{n \setminus i})$ in (\ref{eq:final_rtme_nm1_true_s})
using the following estimate:
\begin{align}
  \widetilde{\mS}(\bar{\alpha};\sX_{n \setminus i})
  & =
  \frac{(1 - \bar{\alpha})p}{n - 1}
  \sum_{\underset{j \neq i}{j = 1}}^n
  \frac{1}{\widetilde{v}_j}\vx_j\vx_j^\top 
  +
  \bar{\alpha}\mT
  ,\, \textrm{\small where }
  \label{eq:approx_scatter_nm1}
  \\
  \widetilde{v}_j
  & = 
  \vx_j^\top \est{\mS}(\bar{\alpha};\sX_n)^{-1} \vx_j 
  ~.
  \nonumber
\end{align}
That is, we plug in the \emph{Regularized} TME $\est{\mS}(\bar{\alpha};\sX_n) \in \SS_p^+$ from 
(\ref{eq:final_rtme_bign}) into equation (\ref{eq:final_rtme_nm1_true_s}) to obtain the new weights 
$\widetilde{v}_j$, for $j \neq i$, $j=1,\dots,n$; then use the new weights $\widetilde{v}_j$ to obtain 
the new estimate $\widetilde{\mS}(\bar{\alpha};\sX_{n \setminus i})$ in (\ref{eq:approx_scatter_nm1}).
Using this approximation, and for a fixed $\bar{\alpha} \in (0,1)$, computing 
$\widetilde{\mS}(\bar{\alpha};\sX_{n \setminus i})$ does not require invoking the RFPI algorithm for 
each $i = 1,\dots,n$.
Instead, the RFPI algorithm will be invoked once to compute $\est{\mS}(\bar{\alpha};\sX_n)$ in 
(\ref{eq:final_rtme_bign}), while $\widetilde{\mS}(\bar{\alpha};\sX_{n \setminus i})$ in (\ref{eq:approx_scatter_nm1}) 
can be directly computed for each $i = 1,\dots,n$.
Using this approximation, the optimal $\alpha$ can now be computed as follows:
\begin{align}
  \est{\alpha}^*_\CV
  & =
  \argmin{\alpha \in (0,1)}
  \quad 
  \widetilde{\sL}_{\CV}(\sX_n,\alpha) ~, \, \textrm{where}
  \label{eq:opt_alpha_approx_loocv}
  \\
  \widetilde{\sL}_{\CV}(\sX_n,\alpha)
  & = 
  \frac{1}{n}
  \sum_{i = 1}^n
  \sL(\vx_i,\widetilde{\mS}(\alpha ; \sX_{n \setminus i})) ~,
  \label{eq:approx_aloocvnll}
\end{align}
and $\widetilde{\sL}_{\CV}(\sX_n,\alpha)$ is the \emph{Approximate} average cross-validated loss; 
we refer to the problem in (\ref{eq:opt_alpha_approx_loocv}) as the \emph{Approximate CVL} (ACVL) method.
For $m$ values of $\alpha$ in $(\alpha_1,\dots,\alpha_m)$, the RFPI algorithm will now consume 
$O(m * T(np^2 + p^3))$ running time from the Approximate CVL method.

\textbf{Remark } In practice, the running time for the ACVL method can be hindered by using grid search 
over large values of $m$ to search for an optimal $\alpha$ in $(\alpha_1,\dots,\alpha_m)$. 
The running time complexity for the ACVL method can be significantly improved if searching for $\alpha$ is done using 
an efficient search technique such as the bisection method.
In this case, the number of iterations $m$ that the bisection method needs to converge to a solution $\alpha^*$ within 
a certain tolerance $\varepsilon$ is upper bounded by $\ceil{\log_2(1/\varepsilon)}$.


\section{Empirical Evaluation}
\label{sec:empirical_evaluation}

In this section, we evaluate our shrinkage coefficient estimation approach on synthetic and real high-dimensional 
datasets, and compare it with other shrinkage coefficient estimation methods in the literature; in particular the 
methods proposed in \cite{chen_wiesel_hero_2011} and \cite{zhang_wiesel_2016}.
Similar to other works in the literature on RTME 
\cite{chen_wiesel_hero_2011,wiesel_2012,pascal_chitour_queck_2014,sun_babu_palomar_2014,ollila_tyler_2014,
goes_lerman_nadler_2020}, we consider the Toeplitz matrix used in \cite{bickel_levina_covreg_thresh_2008} 
to be the population scatter matrix $\mS$ for the elliptical RV in (\ref{eq:gen_model_elliptical_rv}); 
i.e. $\mS = (s_{i,j}) = \gamma^{|i - j|}$, where $\gamma = \{0.1, 0.5, 0.85\}$.
Note that $\mS$ approaches a singular matrix when $\gamma \rightarrow 1$, and $\mS$ approaches the identity matrix
when $\gamma \rightarrow 0$.

\begin{figure*}[t]
\centering
\makebox[\textwidth][c]{\includegraphics[trim=3 0 5 3,clip,width=0.9\textwidth]{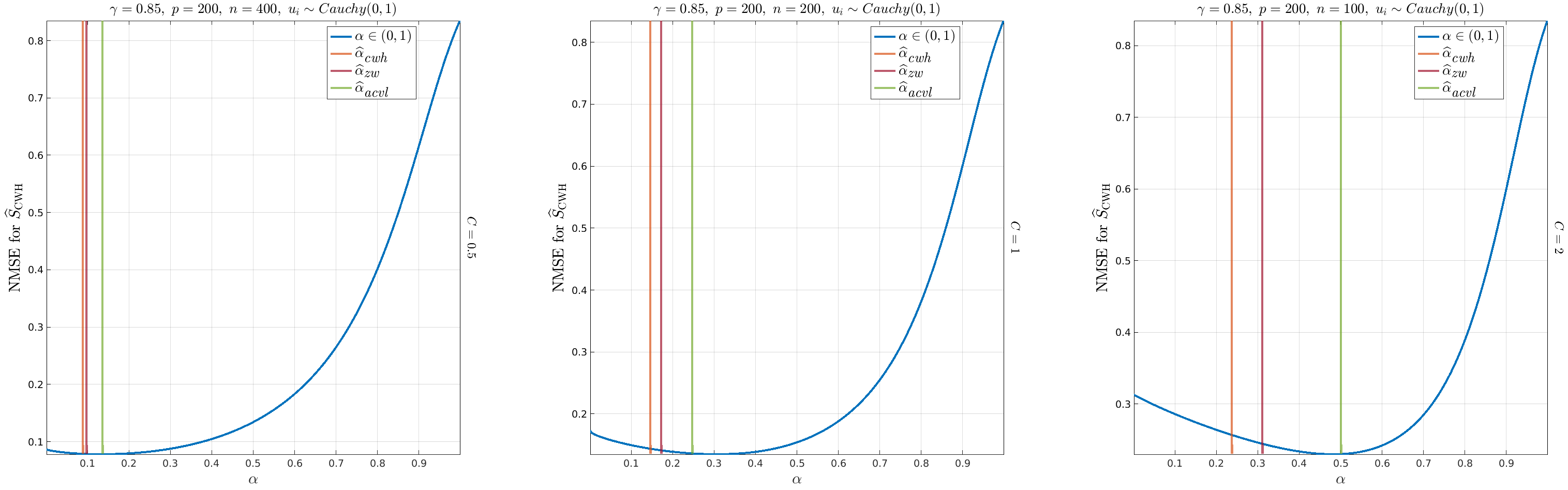}}
\caption{The solid blue line shows the NMSE between the population matrix $\mS$ and the scatter matrix 
$\est{\mS}$ estimated using SBP's RFPI algorithm for $\alpha \in (0,1)$ and $p=200$, in three different 
settings: $p < n$ (left), $p = n$ (middle), and $p > n$ (right). 
The orange, red, and green solid vertical lines indicate the values for shrinkage coefficients 
$\est{\alpha}_\smallCWH$, $\est{\alpha}_\smallZW$, and $\est{\alpha}_\smallACVL$ obtained using the methods 
in \cite[Eq. 13]{chen_wiesel_hero_2011}, \cite[Eq. 12]{zhang_wiesel_2016}, and the Approximate CVL method, respectively.}
\label{fig:exp08_compare_alpha_cwh_zw_aloocv_cauchy_gamma085}
\end{figure*}

The random quantities $u$ and $\vy$ in (\ref{eq:gen_model_elliptical_rv}) are stochastically independent. 
We let $\vy_1,\dots,\vy_n$ be samples from a $p$-variate standard Gaussian distribution $N(\vect{0},\mI)$. 
For r.v. $u$, we consider four different choices for heavy-tailed distributions:
(\emph{i}) $u_i = 1$, which makes $\{\vz_1,\dots,\vz_n\}$ are \iid samples from $N(\vect{0},\mS)$;
(\emph{ii}) $u_i = \sqrt{d/\chi^2_d} $, a Student-T distribution with degrees of freedom $d = 3$;
(\emph{iii}) $u_i = \mathrm{Laplace(0,1)} $, a heavy-tailed distribution with finite moments; and 
(\emph{iv}) $u_i = \mathrm{Cauchy(0,1)} $, a heavy-tailed distribution with undefined moments.
Note that since TME and RTME operate on the normalized samples $\vx_i$, the scalars $u_i$'s cancel out, 
and the resulting plots become identical regardless of the distribution of $u_i$.
For this reason, we show here the plots for the multivariate Cauchy distribution and the multivariate Gaussian 
distributions.

\begin{figure*}
\centering
\makebox[\textwidth][c]{\includegraphics[trim=3 0 5 1,clip,width=0.9\textwidth]{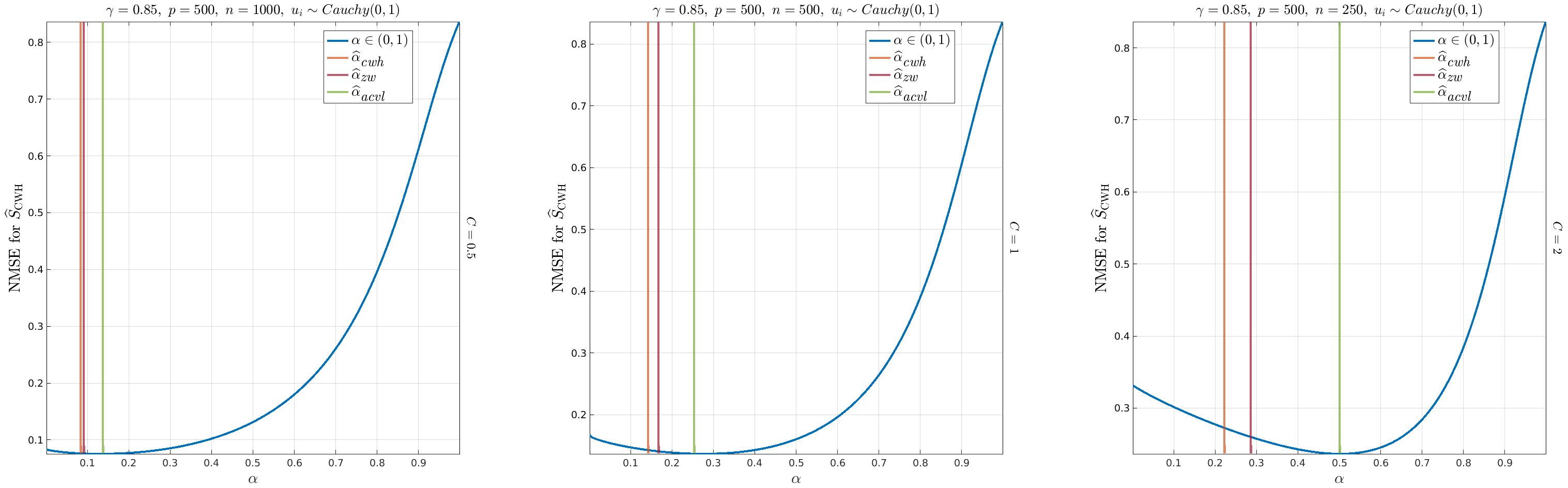}}
\caption{The solid blue line shows the NMSE between the population matrix $\mS$ and the scatter matrix 
$\est{\mS}$ estimated using SBP's RFPI algorithm for $\alpha \in (0,1)$ and $p=500$, in three different 
settings: $p < n$ (left), $p = n$ (middle), and $p > n$ (right). 
The orange, red, and green solid vertical lines indicate the values for shrinkage coefficients 
$\est{\alpha}_\smallCWH$, $\est{\alpha}_\smallZW$, and $\est{\alpha}_\smallACVL$ obtained using the methods 
in \cite[Eq. 13]{chen_wiesel_hero_2011}, \cite[Eq. 12]{zhang_wiesel_2016}, and the Approximate CVL method, respectively.}
\label{fig:exp08_compare_alpha_cwh_zw_aloocv_cauchy_gamma085_p500}
\end{figure*}

The accuracy of an estimator $\est{\mS}$ is measured using the normalized mean squared error (NMSE) 
$\|\est{\mS} - \mS\|_F^2/\|\mS\|_F^2$. 
The convergence criterion for all RFPI algorithms is 
$\|\est{\mS} - \mS\|_F^2 < \epsilon$, where $\epsilon = 1.0e-9$ is the desired solution accuracy.
For Figure (\ref{fig:exp08_compare_alpha_cwh_zw_aloocv_cauchy_gamma085}), $p$ is set to $200$, 
while $n$ is set to three different values $\{400,200,100\}$ to consider three different scenarios: 
$p < n$, $p = n$, and $p > n$, respectively.
For Figure (\ref{fig:exp08_compare_alpha_cwh_zw_aloocv_cauchy_gamma085_p500}), $p$ is set to $500$ while $n$ 
is set to $\{1000,500,250\}$.
The value of $C$ that appears on the right $y$-axis in all Figures is for the ratio $p/n$.
 
Figures (\ref{fig:exp07_optalpha_exact_vs_aloocv_cauchy}) and (\ref{fig:exp07_optalpha_exact_vs_aloocv_gaussian}) 
compare the \emph{Exact} CV loss to the \emph{Approximate} CV loss developed in the previous section, for two 
different elliptical distributions, the multivariate Cauchy distribution (which has undefined moments) and the 
multivariate Gaussian distribution, respectively. 
It can be seen that the Exact CV loss in (\ref{eq:aloocvnll}) (solid blue line) and the Approximate CV loss in 
(\ref{eq:approx_aloocvnll}) (solid red line) are \emph{almost identical} in all settings: $p < n$, $p = n$, $p > n$,
and for all values of $\gamma=\{0.1, 0.5, 0.85\}$ for both distributions. 
This negligible difference between the exact and approximate CV loss supports our proposal that the later can be 
leveraged to estimate a near-optimal value for the shrinkage coefficient $\alpha$.
This can be confirmed by noticing that the optimal $\alpha$ estimated using the Approximate CVL (red square) is 
reasonably close to, or overlaps, the optimal $\alpha$ estimated using the Exact CVL (blue circle) in all nine 
settings for the Cauchy distribution and the Gaussian distribution.
In terms of running time, the legends in Figures (\ref{fig:exp07_optalpha_exact_vs_aloocv_cauchy}) and 
(\ref{fig:exp07_optalpha_exact_vs_aloocv_gaussian}) show the running time (in seconds) for the Exact and Approximate 
CVL methods to estimate $\est{\alpha}^*$, while Table (\ref{tab:speedup_synthetic_data}) shows the corresponding speedup 
for the Approximate CVL method over the Exact CVL method.
We note that the Approximate CVL method is $25\times$ faster (on average) than the Exact CVL method in all the 
different settings for $\gamma$, $p$, and $n$.

\begin{table}[t]
  \centering
  \caption{Speedup for the \emph{Approximate} CVL method over the \emph{Exact} CVL method for each sub-figure in 
  Figures (\ref{fig:exp07_optalpha_exact_vs_aloocv_cauchy}) and (\ref{fig:exp07_optalpha_exact_vs_aloocv_gaussian}).}
  \label{tab:speedup_synthetic_data}
  \begin{tabular}{llccc}
    \toprule 
    Figure & $u_i$ Distribution & \multicolumn{3}{c}{Speedup} \\
    \midrule 
    &                                         & 24.3$\times$ & 35.8$\times$ & 20.6$\times$ \\
    Fig. 2. & $u_i \sim \textit{Cauchy}(0,1)$ & 24.0$\times$ & 33.7$\times$ & 19.0$\times$ \\
    &                                         & 25.0$\times$ & 28.6$\times$ & 18.7$\times$ \\
    \midrule
    &                                         & 23.6$\times$ & 35.2$\times$ & 20.2$\times$ \\
    Fig. 3.  & $u_i \sim N(0,1)$              & 23.8$\times$ & 33.3$\times$ & 19.2$\times$ \\
    &                                         & 25.0$\times$ & 28.5$\times$ & 19.0$\times$ \\
    \bottomrule
  \end{tabular}
\end{table}

Figures (\ref{fig:exp08_compare_alpha_cwh_zw_aloocv_cauchy_gamma085}) and (\ref{fig:exp08_compare_alpha_cwh_zw_aloocv_cauchy_gamma085_p500}) 
compare the shrinkage coefficient estimated using the Approximate CVL method in (\ref{eq:opt_alpha_approx_loocv}), 
denoted by $\est{\alpha}_\smallACVL$, with the shrinkage coefficients estimated from the closed-form expressions 
in \cite[Equation 13]{chen_wiesel_hero_2011}, denoted by $\est{\alpha}_\smallCWH$, 
and
\cite[Equation 12]{zhang_wiesel_2016}, denoted by $\est{\alpha}_\smallZW$.
Although the methods in \cite{chen_wiesel_hero_2011} and \cite{zhang_wiesel_2016} are much faster than the Approximate 
CVL method due to their closed-form expressions, it can be seen that the Approximate CVL method provides more 
accurate estimates for $\alpha$ especially when $p \geq n$.
Also, it can be noticed that the $\alpha$ estimates from \cite{chen_wiesel_hero_2011} and \cite{zhang_wiesel_2016} tend to 
\emph{diverge} from the optimal value as $p$ is growing greater than $n$.
A similar behavior was noticed when using the method in \cite{ollila_tyler_2014} which is also based on asymptotic analysis.
The tendency for methods based on asymptotic analysis and RMT results to over/under estimate the value for $\alpha$ 
is understandable since such methods are based on asymptotic analysis, and make explicit assumptions about the data's 
underlying distribution.
This over/under estimation of $\alpha$ leads to larger values of the NMSE as shown in Figures 
(\ref{fig:exp08_compare_alpha_cwh_zw_aloocv_cauchy_gamma085}) \& 
(\ref{fig:exp08_compare_alpha_cwh_zw_aloocv_cauchy_gamma085_p500}), 
as well as larger values for the LOOCV NLL loss as demonstrated in the following experiments.

\begin{table}[t]
  \centering
  \caption{Comparison results for the first 6 (out of 38) classes from the Extended Yale B dataset for face recognition;
  $n = 64$, $p = 1024$.
  Columns 2, 3, 4, and 5, show the LOOCV NLL loss for scatter matrices estimated using 
  LW \cite{ledoit_wolf_2004}, CWH \cite{chen_wiesel_hero_2011}, ZW \cite{zhang_wiesel_2016}, and the ACVL method, respectively.}
  \label{tab:loocv_loss_faces_yaleb}
  \begin{tabular}{ccccc}
    \toprule 
    Class ID & LW & CWH & ZW & ACVL \\
    \midrule 
    1 & 5677 & 5371 & 5643 & \bfseries 3705 \\
    2	& 5613 & 5440	& 5598 & \bfseries 3706 \\
    3	& 5768 & 5470	& 5749 & \bfseries 3826 \\
    4	& 5403 & 5080	& 5362 & \bfseries 3455 \\
    5	& 5824 & 5435	& 5786 & \bfseries 3716 \\
    6	& 5797 & 5460	& 5761 & \bfseries 3790 \\
    \bottomrule 
  \end{tabular}
\end{table}

Tables (\ref{tab:loocv_loss_faces_yaleb} -- \ref{tab:loocv_loss_uspsdigits}) compare the LOOCV NLL loss for the scatter 
matrices estimated using Ledoit--Wolf (LW) linear shrinkage estimator \cite{ledoit_wolf_2004}, 
and the RTME with shrinkage coefficients from \cite{chen_wiesel_hero_2011}, \cite{zhang_wiesel_2016}, 
and the Approximate CVL method in (\ref{eq:opt_alpha_approx_loocv}).
The comparison between the different estimators was carried out using four real high-dimensional datasets:
(\emph{i}) Images for the first six (6) subjects from the Extended Yale B dataset for face 
recognition\cite{dbfacesyaleab}\footnote{\url{http://cvc.cs.yale.edu/cvc/projects/yalefaces/yalefaces.html}};
(\emph{ii}) Images for the first six (6) object categories from the test set for the CIFAR100 dataset for object 
recognition\footnote{\url{https://www.cs.toronto.edu/~kriz/cifar.html}};
(\emph{iii}) Images for the first six (6) object categories from the test set for the CIFAR10 dataset for object 
recognition; and
(\emph{iv}) Images for the first six (6) digits' classes (0, 1, 2, 3, 4, 5) from the United States Postal Service 
(USPS) dataset for handwritten digits \cite{dbuci}.

The Extended Yale B dataset consists of 2414 frontal-face grayscale (intensity) images for 38 subjects 
-- approx. 64 images per subject -- where each image size (height $\times$ width) is 192 $\times$ 168 pixels.
The images were captured under different poses, lighting conditions, and facial expressions.
The exact face images are cropped and scaled to 32 $\times$ 32 pixels (i.e. $p = 1024$). 
The CIFAR10 and CIFAR100 datasets consist of colored (RGB) images for ten (10) and one hundred (100) objects, 
respectively, from different visual categories (trucks, ships, dogs, mountains, frogs, apples, roads, etc.)
Each colored image has a size of (height $\times$ width $\times$ channels) 32 $\times$ 32 $\times$ 3 which is 
then converted to a grayscale (intensity) image with a final size of 32 $\times$ 32 pixels (i.e. $p = 1024$).%
\footnote{See Matlab's \texttt{rgb2gray()} function for more details.}
The USPS dataset consists of 9298 grayscale images each with a size of 16 $\times$ 16 pixels (i.e. $p = 256$).
The images are obtained from scanning handwritten numerals from envelopes by the U.S. Postal Service and 
they reflect a wide range of handwriting styles.
For all datasets, the data points from each class were centered to have zero mean.

First, from Tables (\ref{tab:loocv_loss_faces_yaleb} -- \ref{tab:loocv_loss_uspsdigits}) it can be seen that 
for most of the cases, scatter matrices estimated using RTME yield lower LOOCV NLL loss  than the scatter matrices 
estimated using LW estimator.
The difference in performance between both classes of estimators is primarily due to the difference in the 
underlying assumption on data distribution; consequently, both classes derive different estimation procedures 
for their respective scatter matrices.
While LW estimator assumes that the data are sampled from a multivariate Gaussian distribution, the class of 
regularized TME assumes that the data are sampled from a multivariate elliptical distribution with heavy tails.
The better performance for RTME suggests that the class of multivariate elliptical distributions can be a better 
alternative than the Gaussian distribution for modeling high-dimensional real data with an (unknown) empirical 
distribution.
Second, in terms of shrinkage coefficients for RTME, it can be seen that the Approximate CVL method yields lower 
LOOCV NLL loss than the methods in \cite{chen_wiesel_hero_2011} and \cite{zhang_wiesel_2016} for all cases in Tables 
(\ref{tab:loocv_loss_faces_yaleb} -- \ref{tab:loocv_loss_uspsdigits}).
This confirms our earlier observation that over/under estimation of the shrinkage coefficient $\alpha$ leads to larger
LOOCV NLL loss which, potentially, may jeopardize the performance of one or more downstream inferential tasks.

\textbf{Discussion} The motivation for the ACVL method is to efficiently estimate an optimal scatter matrix $\mS$,
in the sense of Equation (\ref{eq:opt_scatter_rtme_nll}), using Regularized TME.
Recall that RTME was first proposed to address the `$p > n$' scenario where the original TME cannot be defined.
However, as shown in Table (\ref{tab:loocv_loss_uspsdigits}) for the USPS dataset, the ACVL method is applicable 
and useful for efficiently estimating a scatter matrix using RTME when $n > p$.
It can be seen from Table (\ref{tab:loocv_loss_uspsdigits}) that for different values of $n$, the ACVL method yields 
the lowest LOOCV NLL loss when compared to the methods proposed in \cite{chen_wiesel_hero_2011} and \cite{zhang_wiesel_2016}.

The applicability of the ACVL method to both scenarios, `$p > n$' and `$n > p$', warrants further discussion for the 
scalability of the ACVL method with respect to $n$ and $p$.
In particular, Table (\ref{tab:time_cost_for_exact_vs_approx_rtme}) compares the required average time (in milliseconds) 
to compute the regularized sample covariance matrix $\est{\mS}_{LW}(\sX_{n \setminus i})$ using LW estimator \cite{ledoit_wolf_2004}, 
the exact estimate $\est{\mS}(\bar{\alpha};\sX_{n \setminus i})$ using CWH and ZW, and
the approximate estimate $\widetilde{\mS}(\bar{\alpha};\sX_{n \setminus i})$ in Equation (\ref{eq:approx_scatter_nm1}), 
for a given coefficient $\bar{\alpha}$, and for different values of $n$ and $p$.
Three different observations can be noted from Table (\ref{tab:time_cost_for_exact_vs_approx_rtme}).
First, for small $n$ and `$p \gg n$', computing the approximate estimate $\widetilde{\mS}(\bar{\alpha};\sX_{n \setminus i})$ is 
slightly faster than computing the exact estimate $\est{\mS}(\bar{\alpha};\sX_{n \setminus i})$ using the method of ZW 
\cite{zhang_wiesel_2016}, but significantly faster than computing the exact estimate using the method of CWH 
\cite{chen_wiesel_hero_2011}.
Recall that the ACVL method computes the RTME $\est{\mS}(\bar{\alpha};\sX_n)$ once (the initial overhead), and then 
uses Equation (\ref{eq:approx_scatter_nm1}) to obtain the approximate estimate 
$\widetilde{\mS}(\bar{\alpha};\sX_{n \setminus i})$, for each $i=1,\dots,n$.
Second, as $n$ is increasing, and `$p > n$', computing the approximate estimate 
$\widetilde{\mS}(\bar{\alpha};\sX_{n \setminus i})$ 
becomes significantly cheaper than computing the exact estimate using the methods of CWH and ZW.
Last, as $n$ is increasing, '$n > p$', and for a fixed $p$, the running time for the ACVL method scales mildly with the 
sample size $n$.

\begin{table}[t]
  \centering
  \caption{Comparison results for the first 6 (out of 20) classes from the test set of the CIFAR100 dataset for object recognition;
  $n = 500$, $p = 1024$.}
  \label{tab:loocv_loss_cifar100}
  \begin{tabular}{ccccc}
    \toprule 
    Class ID & LW & CWH & ZW & ACVL \\
    \midrule 
    0 & 849  & 866	 & \bfseries 846.3 & 846.5 \\
    1	& 807	 & 810	 & 799	& \bfseries 767 \\
    2	& 968	 & 984	 & 967	& \bfseries 932 \\
    3	& 810	 & 794	 & 803	& \bfseries 769 \\
    4	& 869	 & 846	 & 859	& \bfseries 812 \\
    5	& 1051 & 1047  & 1043	& \bfseries 1008 \\
    \bottomrule 
  \end{tabular}
\end{table}

\begin{table}[t]
  \centering
  \caption{Comparison results for the first 6 (out of 10) classes from the test set of the CIFAR10 dataset for object recognition;
  $n = 1000$, $p = 1024$.}
  \label{tab:loocv_loss_cifar10}
  \begin{tabular}{ccccc}
    \toprule 
    Class ID & LW & CWH & ZW & ACVL \\
    \midrule 
    0 & 631 & 593	& 612 & \bfseries 590 \\
    1	& 913	& 900	& 906	& \bfseries 894 \\
    2	& 727	& 705	& 718	& \bfseries 694 \\
    3	& 773	& 757	& 772	& \bfseries 755 \\
    4	& 769	& 753	& 761	& \bfseries 739 \\
    5	& 721	& 702	& 719	& \bfseries 699 \\
    \bottomrule 
  \end{tabular}
\end{table}

\begin{table}[t]
  \centering
  \caption{Comparison results for the first 6 (out of 10) classes from the USPS dataset for handwritten digits' recognition; $p=256$.
  Note that the number of samples $n$ varies for each digit's class.}
  \label{tab:loocv_loss_uspsdigits}
  \begin{tabular}{cccccc}
    \toprule 
    Class ID & $n$ & LW & CWH & ZW & ACVL \\
    \midrule 
    0 & 1585 & 268  & 259 &  261 & \bfseries 239  \\
    1	& 1330 &-269	&-374	& -309 & \bfseries -475 \\
    2	& 952	 & 370	& 366	&  369 & \bfseries 342  \\
    3	& 807	 & 336	& 327	&  330 & \bfseries 298  \\
    4	& 795	 & 310	& 293	&  301 & \bfseries 249  \\
    5	& 659	 & 360	& 357	&  359 & \bfseries 337 \\
    \bottomrule 
  \end{tabular}
\end{table}

\setlength{\tabcolsep}{3pt}
\begin{table}[t]
  \centering
  \caption{Average time (in milliseconds) to compute the regularized sample covariance estimate 
  $\est{\mS}_{LW}(\sX_{n \setminus i})$ using LW estimator \cite{ledoit_wolf_2004}, 
  the exact estimate $\est{\mS}(\bar{\alpha};\sX_{n \setminus i})$ using CWH \cite{chen_wiesel_hero_2011} and ZW \cite{zhang_wiesel_2016}, 
  and the approximate estimate $\widetilde{\mS}(\bar{\alpha};\sX_{n \setminus i})$ in Equation (\ref{eq:approx_scatter_nm1}) for a given 
  coefficient $\bar{\alpha}$.}
  \label{tab:time_cost_for_exact_vs_approx_rtme}
  \begin{tabular}{lc cc cccc}
    \toprule 
    Dataset  & ClassID  & $n$  & $p$   & LW & CWH & ZW & ACVL \\
    \midrule 
    Yale B   & 1--6     & 64   & 1024  &  30.8  & 5146.72 & 92.05 & 84.81  \\
    \midrule
    CIFAR100 & 1--6     & 500  & 1024  &  21.31  & 1415.32 & 315.8 & 19.8 \\
    \midrule
    CIFAR10  & 1--6     & 1000 & 1024  &  26.86  & 1020.12 & 564.26 & 25.2 \\
    \midrule 
    USPS     & 0        & 1585 & 256   &  3.88  & 69.27 & 52.33 & 2.81 \\
             & 1        & 1330 & 256   &  3.63   & 89.86 & 53.95 & 2.33 \\
             & 2        & 952  & 256   &  3.27  & 43.83 & 32.89 & 1.87 \\
             & 3        & 807  & 256   &  2.74  & 44.52 & 29.83 & 1.77 \\
             & 4        & 795  & 256   &  2.29  & 55.28 & 31.76 & 1.62 \\
             & 5        & 650  & 256   &  1.57  & 43.73 & 25.54 & 1.54 \\ 
    \bottomrule 
  \end{tabular}
\end{table}


\section{Discussion and Concluding Remarks}
\label{sec:conclusion}

Robust estimation of a high-dimensional covariance matrix from empirical data is well-known to be a challenging task 
in general, and is more daunting when $p \geq n$. 
In this work, we considered TME which is known to be an accurate and efficient robust estimator for the scatter matrix 
when the data are samples from an elliptical distribution with heavy-tails, and $n \gg p$. 
Since TME is not defined when $p \geq n$, various researchers proposed different regularized versions of TME 
where the performance of such estimators depends on a carefully chosen regularization parameter $\alpha \in (0,1)$ 
\cite{abramovich_spencer_2007,chen_wiesel_hero_2011,wiesel_2012,pascal_chitour_queck_2014,sun_babu_palomar_2014,ollila_tyler_2014,zhang_wiesel_2016}.

The research work presented here complements previous efforts in this direction but considers an alternate approach 
for estimating an optimal $\alpha$ for RTME. 
Our approach leverages the given finite sample of high-dimensional points, as well as efficient computation, to 
estimate a near-optimal $\alpha$ for RTME. 
The main driver for the efficient computation is the Approximate LOOCV NLL loss for the estimated scatter matrix with 
respect to parameter $\alpha$ in Equation (\ref{eq:opt_alpha_approx_loocv}).
The resulting procedure, namely the ACVL method, showed positive and promising results in experiments using 
high-dimensional synthetic and real-world data.

The asymptotic properties of LOOCV make an implicit assumption that the estimator enjoys a certain notion 
of algorithmic stability; specifically, that the estimator for scatter matrix $\mS$ is not too sensitive to the removal 
of one sample from $\sX_n$, and hence $\est{\mS}(\bar{\alpha} ; \sX_n) \approx \est{\mS}(\bar{\alpha} ; \sX_{n-1})$. 
$k$-folds cross-validation (KFCV) is another popular technique for model selection that is computationally more
efficient than LOOCV, but also less accurate than LOOCV. 
KFCV may seem a potential candidate to replace the LOOCV in our proposed learning framework. 
Unfortunately, from a stability standpoint, KFCV will require a more stringent stability assumption for the 
estimator of scatter matrix $\mS$ \cite{karim_csaba_apriori_exptailbound_17}; specifically, that the estimator for $\mS$ is not 
too sensitive to the removal of $m = n/k$ samples from $\sX_n$, where $k > 1$ is the number of folds used for KFCV. 
Whether the RFPI algorithm in (\ref{eq:tme_fixed_point_algo}), or any other algorithm for RTME, enjoys such 
a strong notion of stability is an open question that is left for future work. 

An interesting question for future research work is whether the proposed approximation can be extended to other 
covariance matrix estimators, and more generally, to regularization and hyperparameters' selection for different 
classes of learning algorithms.
Another research direction can explore further approximations for the LOOCV loss such that the approximation 
can better exploit the specific structure of the learning algorithm; e.g. algorithms for subspace learning, and 
algorithms for learning mixture models.


\bibliographystyle{IEEEtran} 
\bibliography{latex_bib_karim2022}


\end{document}